\documentclass[letterpaper, 10 pt, conference]{ieeeconf}
\IEEEoverridecommandlockouts
\usepackage{amsmath}
\usepackage{mathtools}
\usepackage{cite}
\usepackage{amsmath,amssymb,amsfonts}
\usepackage[font=small, labelfont=bf]{caption}

\usepackage[ruled, lined, linesnumbered, commentsnumbered, longend, noend]{algorithm2e}
\usepackage{algorithmic}

\usepackage{hyperref}
\usepackage{multicol}
\usepackage{multirow}
\usepackage{makecell}
\usepackage{array}
\usepackage{graphicx}
\usepackage{textcomp}
\usepackage{xcolor}
\usepackage{breqn}
\usepackage{gensymb}
\usepackage{cuted}
\usepackage{capt-of}
\usepackage{times}
\usepackage{colortbl}
\usepackage{xcolor}
\usepackage{pgfplots}
\usepackage{dblfloatfix}
\usepackage{booktabs}
\usepackage{diagbox}

\newcommand{\dgmask}{{\fontfamily{lmtt}\selectfont DGMask}}

\usepackage{fancyhdr} 
\renewcommand{\arraystretch}{1.1}

\title{\LARGE \bf \textit{DeGuV}: Depth-Guided Visual Reinforcement Learning for Generalization and Interpretability in Manipulation}

\author{Tien Pham$^{1}$, Xinyun Chi$^{1}$, Khang Nguyen$^{2}$, Manfred Huber$^{2}$, and Angelo Cangelosi$^{1}$ {
\thanks{*This work has been partially supported by Horizon Europe (and UKRI Horizon Guaranteed Fund)  under the Marie Skłodowska-Curie grant agreement No 101072488 (TRAIL). This work was also in part supported by a project funded by the EPSRC Prosperity grant CRADLE (EP/X02489X/1) and by the Air Force Office of Scientific Research, USAF, under the CASPER++ Awards (FA8655-24-1-7047).} %
\footnotesize \thanks{$^{1}$Cognitive Robotics Lab, School of Computer Science, University of Manchester, UK; $^{2}$Learning and Adaptive Robotics Lab, Department of Computer Science and Engineering, University of Texas at Arlington, USA; (corresponding email: \href{mailto:canhantien.pham@manchester.ac.uk}{\text{canhantien.pham@manchester.ac.uk}}).}}}

\begin{document}

\maketitle
\thispagestyle{empty}
\pagestyle{empty}

\begin{strip}
    \vspace{-70pt}
    \centering
    \includegraphics[width=1.00\linewidth]{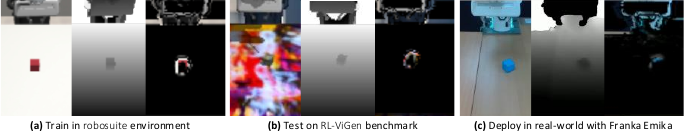}
    \vspace{-12pt}
    \captionof{figure}{\textbf{(a)} \textit{DeGuV} is trained with robotic manipulation tasks in \texttt{robosuite} environment, \textbf{(b)} is evaluated on the \texttt{RL-ViGen} benchmark for generalization and interpretability verfication, and \textbf{(c)} is deployed on a Franka Emika robot to demonstrate its zero-shot sim-to-real transferability. The RGB-D observations and the masked image produced by \textit{DeGuV} are presented from left to right in each case.}
    \vspace{-12pt}
    \label{fig:overview_method}
\end{strip}

\begin{abstract}
    Reinforcement learning (RL) agents can learn to solve complex tasks from visual inputs, but generalizing these learned skills to new environments remains a major challenge in RL application, especially robotics. While data augmentation can improve generalization, it often compromises sample efficiency and training stability. This paper introduces \textit{DeGuV}, an RL framework that enhances both generalization and sample efficiency. In specific, we leverage a learnable \textit{masker network} that produces a mask from the depth input, preserving only critical visual information while discarding irrelevant pixels. Through this, we ensure that our RL agents focus on essential features, improving robustness under data augmentation. In addition, we incorporate contrastive learning and stabilize Q-value estimation under augmentation to further enhance sample efficiency and training stability. We evaluate our proposed method on the RL-ViGen benchmark using the Franka Emika robot and demonstrate its effectiveness in zero-shot sim-to-real transfer. Our results show that \textit{DeGuV} outperforms state-of-the-art methods in both generalization and sample efficiency while also improving interpretability by highlighting the most relevant regions in the visual input. Our implementation is available at: \href{https://github.com/tiencapham/DeGuV}{\texttt{https://github.com/tiencapham/DeGuV}}.
\end{abstract}

\section{Introduction}
\label{sec:introduction}


Vision-based reinforcement learning (RL) has demonstrated its impact in various applications, including video games \cite{kapturowski2022human, schmidt2021fast, badia2020agent57}, autonomous navigation \cite{mirowski2016learning, zhu2017target, zhang2021end}, and robotics \cite{akkaya2019solving, haarnoja2024learning, jangir2022look}. As RL agents often learn from simulation environments \cite{zhang2018study}, they are precluded from generalizing unseen environments in which visual observations differ from those encountered during training. The distribution shift, caused by variations in object textures, colors, and environmental radiance (Fig. \ref{fig:overview_method}), significantly degrades real-world performance. To better generalize these factors, many researchers have explored data augmentation techniques \cite{grooten2024madi, hansen2021stabilizing, ha2023dream, bertoin2022look, yuan2024learning, raileanu2021automatic, hansen2021generalization, liang2024visarl} to diversify training data. Despite enhancing robustness, these methods raise sample efficiency concerns \cite{laskin2020reinforcement, kostrikov2020image} and destabilize training processes \cite{hansen2021stabilizing} of visual RL models for robotics.   

\begin{figure*}[t]
    \centering
    \vspace{4pt}
    \includegraphics[width=1.00\linewidth]{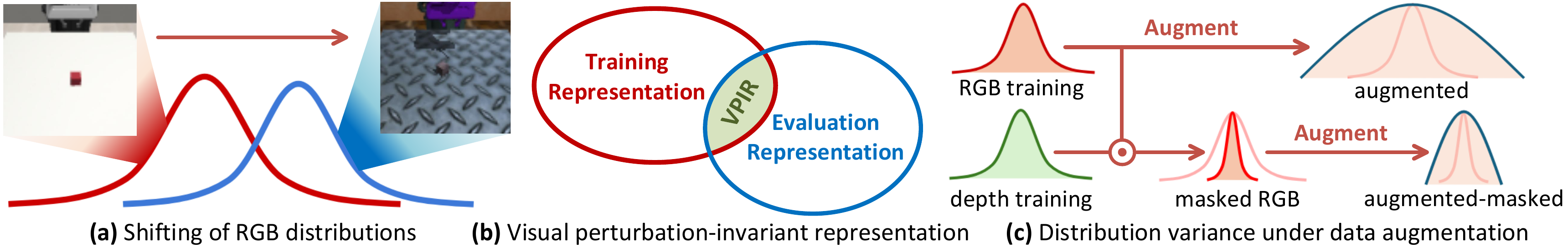}
    \caption{Illustrations of \textbf{(a)} generalization problem in visual RL posed by distribution shifting, \textbf{(b)} visual perturbation-invariant space between training and evaluation representations, \textbf{(c)} changes of variances of RGB and depth distributions under data augmentation.}
    \label{fig:masked_distribution}
    \vspace{-17pt}
\end{figure*}


To address these concerns, we present \textit{DeGuV}, an RL framework that improves generalization while maintaining sample efficiency. Our approach leverages depth input to mitigate the intrinsic variations caused by data augmentation. We ensure that RL agents focus only on the most informative regions by employing a depth-guided masking module to filter out task-irrelevant pixels in RGB images. Through this, we are able to reduce unnecessary variations in augmented training data, enhancing sample efficiency while minimizing the gap between training and evaluation distributions. To further reinforce visual perturbation invariance, we incorporate contrastive learning to facilitate the model to focus on consistent, task-relevant features across varying visual conditions. Lastly, we combine Stabilized Q-Value Estimation (SVEA) \cite{hansen2021stabilizing} to improve the training stability of \textit{DeGuV}.


We evaluate the proposed method on four tasks and three evaluation modalities in the RL-ViGen benchmark \cite{yuan2024rl}. Our results show that we achieve superior results compared to state-of-the-art methods in generalization while maintaining sample efficiency. Moreover, our model can enhance interpretability by visualizing attention through its learned masks, providing insight into the agent’s decision-making process. For real-robot experiments, we illustrate the feasibility of sim-to-real transfer in a zero-shot manner. 


Our contributions are summarized as follows:
\begin{itemize}
    \item We formulate the generalization problem inherent in visual RL under data augmentation, solving the predicament between the model's generalization capability and learning complexity.
    \item We introduce an RL framework, leveraging depth input to reduce visual observation's variation internally during training and evaluation, enhancing the trained policy's generalization capability and sample efficiency.
    \item We conduct experiments on the generalization benchmark against state-of-the-art techniques. The results show improvements in generalization ability, sample efficiency, and interpretability.
    \vspace{-6pt}
\end{itemize}. 
\vspace{-8pt}

\section{Related Work}

\textbf{Generalization in Reinforcement Learning}: Prior works have been investigated in enhancing generalization capabilities of RL using various techniques such as contrastive learning \cite{agarwal2021contrastive, laskin2020curl}, data augmentation \cite{grooten2024madi, hansen2021stabilizing, ha2023dream, bertoin2022look, yuan2024learning, raileanu2021automatic, hansen2021generalization, liang2024visarl, xie2024decomposing}, domain adaptation \cite{xing2021domain}, and incorporation of pre-trained image encoders \cite{yuan2022pre}. While data augmentation and other techniques have shown effectiveness in improving the generalization ability of RL agents, they usually introduce additional complexity in optimal policy learning \cite{hansen2021stabilizing}. Chen \textit{et al.} \cite{chen2024focus} introduced the Focus-then-Decide framework with SAM \cite{kirillov2023segany} to assist the agent in training efficiently under data augmentation. However, SAM requires an additional computing workload for both the training and inference processes, which worsens the computational load of RL models, especially for robotic applications. Yuan \textit{et al.} \cite{yuan2024learning} uses multi-view representation learning to endow the generalization ability of agents under various types of data augmentation. In this work, we aim to stabilize and improve RL agents' performance by intrinsically reducing the variation in the augmented observation distribution.

\textbf{Interpretability by Visual Masking:} A few studies focus on enhancing RL agents' generalization ability and interpretability by selectively masking input portions. Yu \textit{et al.} \cite{yu2022mask} randomly obscure portions of the input and apply an auxiliary loss to recover the missing pixels. Bertoin \textit{et al.} \cite{bertoin2022look} proposes a Saliency-Guided Q-Network (SGQN) that self-supervisedly learns to generate the mask based on the gradient obtained during the training. SGQN is sensitive to hyperparameter $\texttt{sgqn\text{-}quantile}$ (often set to $95\% \text{-} 98\%$), which determines how many pixels are masked. Tomar \textit{et al.} \cite{tomar2024ignorance} introduce InfoGating with U-Net to create the mask from the downstream loss. Grooten \textit{et al.} \cite{grooten2024madi} present MaDi, which generates the soft mask solely from the reward signal of the environment. However, the mask generated by InfoGating and MaDi depends on the perturbed observations. Furthermore, all the mentioned works use augmented visual observations to produce the mask, while we use depth input -- a more stable distribution under augmentation.

\textbf{RGB-D Fusion for Visual Reinforcement Learning:} While most image-based RL studies focus on visual observations, some works incorporate visual and depth inputs to enrich environmental information and improve the performance of RL agents. The two data streams could be encoded separately and then fused by simple concatenation \cite{balakrishnan2021curriculum, joshi2020robotic, friji2020dqn} or by a cross-attention mechanism \cite{james2022q}. Balakrishnan \textit{et al.} \cite{balakrishnan2021curriculum} uses a pre-trained Twin Variational Autoencoder to extract the environment's latent embedding simultaneously, which is used to train RL policies. James \textit{et al.} \cite{james2022q} introduces a Q-Attention Agent that is used to extract cross-attention features, combined with a Next-Best Pose Agent to predict the next-best poses from RGB and point cloud inputs. Again, within this work, we only use depth images to produce the masks, which helps to reduce the variation in perturbed visual observations. The primary semantic information is still extracted from RGB observations.

\section{Problem Formulation}
\label{subsec:problem_formulation}

Learning of generalizable policies in a Markov Decision Process (MDP) is formulated as an invariant representation learning problem, where learned policies could maintain their performance in unseen environments shifted from the visual training distribution. The interaction between environment and policy is constructed as an MDP: $\mathcal{M} = \langle \mathcal{S}, \mathcal{A}, \mathcal{P}, r, \gamma \rangle$, with $\mathcal{S}$ is the state space, $\mathcal{A}$ is the action space, $\mathcal{P}: \mathcal{S} \times \mathcal{A} \rightarrow \mathcal{S}$ is the transition function, $r: \mathcal{S} \times \mathcal{A} \rightarrow \mathbf{R}$ is the reward function, and $\gamma$ is the discount factor. 

\begin{figure*}
    \centering
    \vspace{4pt}
    \includegraphics[width=0.85\linewidth]{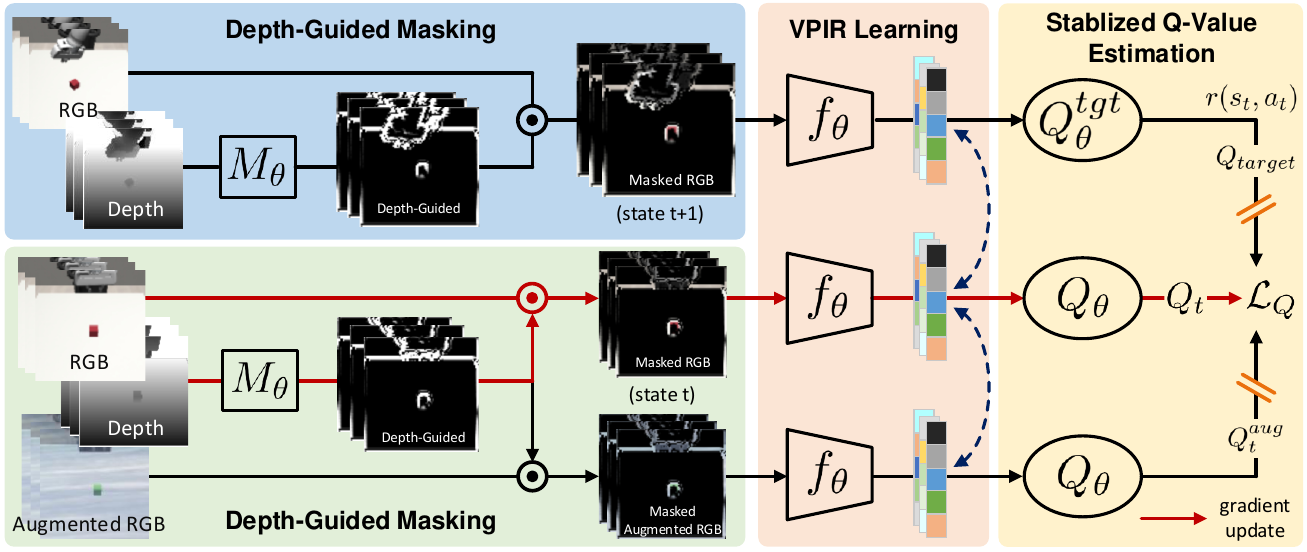}
    \caption{The training pipeline of \textit{DeGuV} consists three main components: Depth-Guided Masking module takes in depth inputs to create the mask that erases non-relevant pixels in RGB image, VPIR Learning module employs contrastive loss in Eq. \ref{equ:infonce} to facilitate the RL agent to extract invariant representations, and Stablized Q-Value Estimation is applied to stabilize the optimal policy training by avoiding non-stationary updates using $\texttt{stop\_grad}$ operators ({\textcolor{orange}{\textbf{//}})}. Details of this training are provided in Alg. \ref{alg:deguv}.}
    \label{fig:architecture_depth_masking}
    \vspace{-17pt}
\end{figure*}

To overcome partial observability problems of static visual input, we wrap a state ${s}_t$ to include $k$ consecutive frames $\{\mathbf{o}_t, \mathbf{o}_{t+1}, \dots, \mathbf{o}_{t+k}\}$, $\mathbf{o}_i \in \mathcal{O}$, where $\mathcal{O}$ is the high-dimensional observation space. 
We let the state space $\mathcal{S}$ is the one constructed as $(P^{\text{RGB}} \cup P^{D})^k$ with $P^{\text{RGB}}$ and $P^{D}$ are RGB and depth image pixel space, respectively, from the observation space $\mathcal{O}$.

Our objective is to learn a policy $\pi: \mathcal{S \rightarrow A}$ that maximizes the discounted return $\mathcal{R}_t=\mathbf{E}_{\Gamma \sim \pi} \left[ \sum_{t=0}^{T} \gamma^t \, r(s_t, a_t) \right]$ along a trajectory $\Gamma = (s_0, s_1, \dots, s_T)$, which are sampled from the observation space following the policy $\pi$ from the initial state $s_0$. The policy $\pi$ is parameterized by a collection of learnable parameters $\theta$, denoted as $\pi_{\theta}$. We aim to learn the parameters $\theta$ so that $\pi_\theta$ generalizes well across unseen visual MDPs, denoted as $\mathcal{\overline{M}} = \langle \mathcal{\overline{S}}, \mathcal{A}, \mathcal{P}, r, \gamma \rangle$, where the states $\overline{s}_t \in \overline{\mathcal{S}}$ are constructed from the observations $\{\overline{o}_t, \overline{o}_{t+1}, \dots, \overline{o}_{t+k} \} \in \mathcal{\overline{O}}$. The perturbed observation space $\mathcal{\overline{O}}$ has a visual distribution shift, such as changes in color and brightness, from the original observation space $\mathcal{O}$.

\section{Methodology}
\label{sec:method}
To learn the optimal policy with the continuous action space in robotic manipulation, we leverage the Soft Actor-Critic (SAC) \cite{haarnoja2018soft} algorithm to approximate the optimal state-action value function $Q^{*}$ using a parameterized critic $Q_{\theta}$. This is achieved by minimizing the Bellman residual \cite{sutton1988learning}:
\begin{equation}
    R = (r(s_t,a_t) + \gamma \max \left( Q^{\text{tgt}}_{\theta} (s_{t+1}, a^{'}_t) \right) - Q_\theta(s_t, a_t))
    \label{equ:bellman}
\end{equation}

The actor, defined by a stochastic policy, $\phi_{\theta}$, optimizes the output of the critic network while ensuring high entropy. SAC often incorporates an optional shared encoder $f_{\theta}$ for image-based environments. The critic and encoder use target networks initialized with the same parameters $\theta_{\text{tgt}} = \theta$, updated via an exponential moving average $\theta_{\text{tgt}} \gets (1 - \tau)\theta_{\text{tgt}} + \tau\theta$. We formulate the optimal policy learning problem as a two-staged problem:
\begin{enumerate}
    \item \textit{Latent Representation Learning:} The agent learns to encode the latent representation $z_t = f_{\theta}(s_t)$ of size $N \ll \text{dim}(\mathcal{S})$ where $f_{\theta}:\mathcal{S} \rightarrow \mathcal{Z}$ is an encoder parameterized by $\theta$ and $\mathcal{Z}$ represents the latent space.
    \item \textit{Decision Making:} The agent learns to choose the optimal action from the latent representation $a_t \sim \phi_{\theta}(a_t \mid z_t)$ to maximize the discounted return.
\end{enumerate}

\subsection{Data Augmentation}
As the well-explored approach in prior works \cite{grooten2024madi, hansen2021stabilizing, bertoin2022look, yuan2024learning, ha2023dream}, data augmentation techniques, where observations are augmented to create a broader visual distribution during network training (Fig. \ref{fig:masked_distribution}c), can prevent the agent from overfitting to the training environment and improve generalization ability. Under data augmentation, the model is incentivized to learn task-relevant semantic information that could generalize well in visual perturbations, ensuring the trained policy's generalization in various unseen scenarios (Fig. \ref{fig:masked_distribution}a and Fig. \ref{fig:masked_distribution}b). This leads to the Visual Perturbation-Invariant Representation's definition (VPIR), formulated as follows:

\textbf{Definition.} \textit{Given an MDP $\mathcal{M}$ that has a state $s \in \mathcal{S}$ and its perturbed counterpart $\overline{s} \in \overline{\mathcal{S}}$, there is a set of latent representations $z^*$ that could be encoded by the optimal policy to represent the task-relevant semantic information from both $s$ and $\overline{s}$. Mathematically:}
\begin{equation}
    \exists z^* \in \mathcal{Z}, z^*=f_\theta(s)=f_\theta(\overline{s}), s \in \mathcal{S}, \overline{s} \in \mathcal{\overline{S}}
\end{equation}

However, the data augmentation phases inevitably increase the state space size, creating additional complexity in the training process and reducing the sample efficiency. We observe that the augmented distribution's variance is larger than the original variance: $\texttt{Var} \left( P^{\text{RGB}}_{\text{aug}} \right) \gg \texttt{Var} \left( P^{\text{RGB}} \right)$.

\subsection{Distractions}
We denote $p^{\text{RGB}}_{\text{distraction}}$ as the unrelated input pixels, the so-called visual distraction, to the task, which could be safely excluded during optimal policy learning. Meanwhile, the remaining pixels are called task-related pixels $P^{\text{RGB}}_{\text{relevant}} = P^{\text{RGB}}_{\text{aug}} \setminus P^{\text{RGB}}_{\text{distraction}}$. Therefore, VPIR is extracted from only the task-related pixels, we derive the relationship between their distribution variance as follows:
\begin{equation}
    \texttt{Var} \left( P^{\text{RGB}}_{\text{aug}} \right) = \texttt{Var} \left( P^{\text{RGB}}_{\text{relevant}} \right) + \texttt{Var} \left( P^{\text{RGB}}_{\text{distraction}} \right)
    \label{eq:var}
\end{equation}

When the distractions are properly removed, the variance created on those pixels will vanish in Eq. \ref{eq:var}, reducing the complexity of optimal policy learning. We divide the \textit{latent representation learning} into \textit{visual distraction masking}, where the agent learns to erase distractions in visual states, and \textit{learning latent representation from masked observations}, where the agent is trained to encode the latent representation from the masked state.

The visual distraction masking problem complements the MDP, identifying the irrelevant pixels to the task and Q-value estimation in the Bellman residual. Since the optimal $Q^*$ remains unchanged by distractions, minimizing the Bellman residual (Eq. \ref{equ:bellman}) simultaneously accomplishes learning of an optimal policy and masking visual distractions. This means the agent is incentivized to ignore distractions and only focus on relevant pixels to achieve an optimal policy.
However, learning to generate the mask from highly varied augmented visual distribution $P^{\text{RGB}}_{\text{aug}}$ is more complex than learning from original visual distribution, which leads to unstable optimization. Meanwhile, the depth distribution is invariant to visual perturbations and remains stable under data augmentation (Fig. \ref{fig:masked_distribution}c), thus could serve as a reliable source for distraction masking procedure.

\setlength{\textfloatsep}{0pt}
\begin{algorithm}[t]
    \caption{DeGuV Training}
    \label{alg:deguv}
    \begin{normalsize}
        \DontPrintSemicolon
        \SetKwInOut{KwIn}{Input}
        \SetKwInOut{KwOut}{Output}
        \SetKwFunction{FTrain}{DeGuV\_train}
        \SetKwProg{Pn}{function}{}{}
        
        \KwIn{$f_{\theta}, Q_{\theta}, \phi_{\theta}, M_{\theta} \coloneqq$ initialized networks \\
              $Q^{tgt}_{\theta}, f^{tgt}_{\theta} \gets Q_{\theta}, f_{\theta} \coloneqq$ target networks \\
              $T \coloneqq$ total timesteps}
        \KwOut{$f_{\theta}, Q_{\theta}, \phi_{\theta}, M_{\theta} \coloneqq $ trained networks}
        
        \Pn{\FTrain{$f_{\theta}, Q_{\theta}, \phi_{\theta}, M_{\theta}, T$}}{
            \For{$t \in [1, \ldots, T]$}{
            
                \small{\texttt{$\triangleright$ ------- act phase -------}} \\
                $s^{\text{masked}}_t \gets$ \dgmask$(s_t)$ \\
                $a_t \sim \phi_{\theta}(\cdot | f_{\theta}(s^{\text{masked}}_t))$ \\
                $s_{t+1}, r_t \sim \mathcal{P}(\cdot | s_t, a_t)$ \\
                $\mathcal{B} \gets \mathcal{B} \cup (s_t, a_t, r_t, s_{t+1})$ \\

                \small{\texttt{$\triangleright$ ------- update phase -------}} \\
                $\{s_b, a_b, r_b, s'_b\} \sim \mathcal{B}$ \\
                $s_b, s'_b \gets$ \dgmask$(s_b),$ \dgmask$(s'_b)$ \\
                $s_b, s'_b \gets [s_b, \tau(s_b, \nu)], [s'_b, \tau(s'_b, \nu')]$ \\
                
                \small{\texttt{$\triangleright$ ------- update actor -------}} \\
                $\theta_{\phi} \gets \theta_{\phi} - \nabla_{\theta_{\phi}} \mathcal{L}_{\phi}(s_b)$ 
                
                \small{\texttt{$\triangleright$ ------- update critics -------}} \\
                \For{$N \in \{Q_{\theta}, f_{\theta}, M_{\theta}\}$}{
                    $\theta_N \gets \theta_N - \nabla_{\theta_N} \mathcal{L}_Q(s_b, a_b, r_b, s'_b)$ 
                }
                \For{$N \in \{Q_{\theta}\}$}{
                    $\theta^{tgt}_N \gets (1 - \tau)\theta^{tgt}_N + \tau \theta_N$
                }

                \small{\texttt{$\triangleright$ ------- update auxiliary -------}} \\
                $\theta_f \gets \theta_f - \nabla_{\theta_f} \mathcal{L}_{\text{InfoNCE}}$
                
            }
            \KwRet{$f_{\theta}, Q_{\theta}, \phi_{\theta}, M_{\theta}$}
        }
    \end{normalsize}
\end{algorithm}

\begin{table*}[t]
    \vspace{5pt}
    \centering
    \caption{Comparisons of episode returns ($\mu, \sigma$) of \textit{DeGuV} against other baselines on RL-Vigen benchmark \cite{yuan2024rl} in four training and evaluation environments: \texttt{train}, \texttt{easy}, \texttt{medium}, and \texttt{hard} of four tasks \texttt{Lift}, \texttt{Door}, \texttt{NutAssemblyRound}, and \texttt{TwoArmPegInHole}.}
    \label{tab:eval_result}
    \vspace{-2pt}
    \renewcommand{\arraystretch}{1.2}
    \begin{tabular}{c c c c c c c | c}
        \toprule
        \multicolumn{2}{c}{\diagbox{Task}{Baseline}} & DrQv2 \cite{yarats2021drqv2} & CURL \cite{agarwal2021contrastive} & SGQN \cite{bertoin2022look} & SVEA \cite{hansen2021stabilizing} & MaDi \cite{grooten2024madi} & DeGuV (ours) \\
        \midrule \midrule
        \multirow{3}{*}{\rotatebox[origin=c]{90}{\parbox{1.5cm}{\centering \texttt{Lift} }}} & \texttt{train} & 242.21 $\pm$ 187.82 & 125.65 $\pm$ 84.10 & 142.86 $\pm$ 43.10 &  234.25 $\pm$ 15.83 & 364.87 $\pm$ 34.66 & \textbf{443.85 $\pm$ 34.26} \\
        & \texttt{easy} & 17.14 $\pm$ 70.08 & 74.16 $\pm$ 71.06 & 57.16 $\pm$ 35.00 & 118.84 $\pm$ 90.42 & 130.47 $\pm$ 131.68 & \textbf{293.80 $\pm$ 160.07} \\
        & \texttt{medium} & 0.11 $\pm$ 0.11 & 9.70 $\pm$ 15.91 & 39.65 $\pm$ 20.92 & 12.74 $\pm$ 23.96 & 10.79 $\pm$ 17.23 & \textbf{324.67 $\pm$ 110.79} \\
        & \texttt{hard} & 0.16 $\pm$ 0.47 & 13.73 $\pm$ 17.14 & 23.92 $\pm$ 15.86 & 12.24 $\pm$ 25.50 & 23.35 $\pm$ 20.56 & \textbf{227.32 $\pm$ 134.84} \\
        \midrule
        \multirow{3}{*}{\rotatebox[origin=c]{90}{\parbox{1.5cm}{\centering \texttt{Door} }}} & \texttt{train} & 387.35 $\pm$ 169.50 &  438.12 $\pm$ 74.14 & 441.61 $\pm$ 137.58 & \textbf{475.82 $\pm$ 27.99} & 444.41 $\pm$ 121.52 & 467.23 $\pm$ 70.82 \\
        & \texttt{easy} & 365.66 $\pm$ 186.72 & 436.39 $\pm$ 78.44 & 440.87 $\pm$ 139.02 & 464.73 $\pm$ 60.45 & 405.72 $\pm$ 145.43 & \textbf{476.47 $\pm$ 47.23} \\
        & \texttt{medium} & 160.44 $\pm$ 215.12 & 190.40 $\pm$ 178.89 & 236.71 $\pm$ 233.97 & 247.97 $\pm$ 217.57 & 204.19 $\pm$ 198.21 & \textbf{333.32 $\pm$ 203.53} \\
        & \texttt{hard} & 113.39 $\pm$ 182.30 & 135.65 $\pm$ 184.39 & 126.73 $\pm$ 204.60 & 256.17 $\pm$ 210.97 & 135.65 $\pm$ 184.39 & \textbf{447.24 $\pm$ 103.55} \\
        \midrule
        \multirow{3}{*}{\rotatebox[origin=c]{90}{\parbox{1.5cm}{\centering \texttt{Nut} \\ \texttt{Assembly} \\ \texttt{Round} }}} & \texttt{train} & 74.72 $\pm$ 60.64  & 61.91 $\pm$ 54.22 & 106.10 $\pm$ 67.89 & 135.01 $\pm$ 63.30 & \textbf{138.55 $\pm$ 59.55} & 137.93 $\pm$ 56.53  \\
        & \texttt{easy} & 50.63 $\pm$ 52.81 & 61.49 $\pm$ 55.95 & 67.23 $\pm$ 68.29 & \textbf{130.98 $\pm$ 67.71} & 127.31 $\pm$ 63.08 & 113.34 $\pm$ 67.50  \\
        & \texttt{medium} & 6.35 $\pm$ 12.28 & 38.91 $\pm$ 39.01 & 14.93 $\pm$ 31.83 & 4.30 $\pm$ 7.32 & 0.71 $\pm$ 1.67& \textbf{105.82 $\pm$ 62.52}  \\
        & \texttt{hard} & 13.62 $\pm$ 27.13 & 22.43 $\pm$ 30.84 & 7.10 $\pm$ 13.87 & 8.08 $\pm$ 14.17 & 0.6 $\pm$ 1.22 & \textbf{104.50 $\pm$ 66.99}  \\
        \midrule
        \multirow{3}{*}{\rotatebox[origin=c]{90}{\parbox{1.5cm}{\centering \texttt{TwoArm} \\ \texttt{PegInHole} }}} & \texttt{train} & 288.37 $\pm$ 20.61 & 314.99 $\pm$ 15.23 & 286.20 $\pm$ 27.59 & \textbf{422.62 $\pm$ 34.07} & 355.65 $\pm$ 5.04 & 381.07 $\pm$ 53.02  \\
        & \texttt{easy} & 284.53 $\pm$ 20.85 & 310.93 $\pm$ 20.63 & 365.97 $\pm$ 72.08 & 365.97 $\pm$ 72.08 & 355.78 $\pm$ 5.55 & \textbf{376.29 $\pm$ 47.29}  \\
        & \texttt{medium} & 190.75 $\pm$ 21.97 & 196.27 $\pm$ 28.92 & 211.96 $\pm$ 40.97 & 126.96 $\pm$ 12.61 & 280.23 $\pm$ 18.42 & \textbf{378.18 $\pm$ 43.28}  \\
        & \texttt{hard} & 177.19 $\pm$ 35.54 & 186.40 $\pm$ 30.17 & 202.85 $\pm$ 45.38 & 123.12 $\pm$ 17.43 & 225.34 $\pm$ 41.66 & \textbf{372.01 $\pm$ 42.27}  \\
        \midrule
        \multicolumn{2}{c}{\textbf{Average}} & 148.29 $\pm$ 169.81 & 163.57 $\pm$ 160.04 & 168.05 $\pm$ 168.19 & 196.24 $\pm$ 180.74 & 200.23 $\pm$ 174.35 & \textbf{311.44 $\pm$ 159.59}\\
        \bottomrule
    \end{tabular}
    \vspace{-16pt}
\end{table*}

\subsection{Depth-Guided Masking}
\label{subsec:depth-guided-masking}
To reduce the variance of observation distribution caused by the data augmentation, we introduce a depth-guided masker $M_{\theta}$ that leverages the depth input from the state to zero out the irrelevant RGB pixels, as stated in the following Eq. \ref{eq:dgmask}. The masker produces a scalar matrix in the $[0, 1]$ range to determine its relevance to the task. As depth input should be invariant to data augmentation, the mask could be applied element-wise for both the original visual input and its augmented counterpart, leaving minimal variance in visual distribution needed for generalization. The depth-guided masking, \dgmask, is formulated as follows:
\begin{subequations}
    \begin{equation}
        \left(s^{\text{RGB}}_t, s^{D}_t\right) \gets s_t
    \end{equation}
    \begin{equation}
        s^{\text{masked}}_t \gets s^{\text{RGB}}_t \odot M_{\theta}(s^{D}_t)
    \end{equation}
    \label{eq:dgmask}
    \vspace{-10pt}
\end{subequations}

The depth-guided masker $M_{\theta}$ is a learnable network composed of convolutional layers with the ReLU activation function in between. The last layer is followed by a $\texttt{Hardtanh}(\cdot)$ activation function to regularize its value in the range of $[0,1]$. Our mask can eliminate irrelevant pixels using $\texttt{Hardtanh}$ rather than merely reducing their values close to zero. As the masker is fully differentiable, it could be trained during the backward step with the critic update. Additionally, the mask generated by $M_{\theta}$ also enhances the interpretability of the model by highlighting the regions of the visual input that are most relevant for task completion.

\subsection{Visual Perturbation-Invariant Representation Learning}
\label{subsec:vpir}
Contrastive learning is used to train the encoder, which extracts VPIR in a self-supervised manner. In this approach, given a query $\mathbf{q}$, the goal is to enhance the similarity between $\mathbf{q}$ and its corresponding positive key $\mathbf{k^+}$, while simultaneously reducing the similarity between $\mathbf{q}$ and each negative key $\mathbf{k^-}$ in a training batch. We quantify the disparities between vectors $\mathbf{q}$ and $\mathbf{k}$ using cosine similarity:
\begin{equation}
    \text{sim}(\mathbf{q},\mathbf{k}) = \frac{\mathbf{q}^T\mathbf{k}}{\|\mathbf{q}\| \| \mathbf{k}\|}
\end{equation}

The InfoNCE loss \cite{oord2018representation}, $\mathcal{L}_{\text{InfoNCE}}$, is used to penalize the model for learning VPIR in contrastive learning with $\tau$ as the temperature hyperparameter by the following expression:
\begin{equation}
    -\log \left[ \frac{\exp\left(\frac{\text{sim}(\mathbf{q}^T,\mathbf{k}^+)}{\tau}\right)}{\exp\left(\frac{\text{sim}(\mathbf{q}^T,\mathbf{k}^+)}{\tau}\right) +  
    \sum_{i=0}^M{\exp\left(\frac{\text{sim}(\mathbf{q}^T,\mathbf{k}^-_i)}{\tau}\right)}} \right],
    \label{equ:infonce}
\end{equation}

Given a masked visual state $s^{\text{masked}}_t$ and its augmented counterpart $s^{\text{masked} + \text{aug}}_t$, an encoder $f_{\theta}$ needs to learn to extract the states into a VPIR $z^*$ so that it can be generalized into the visual perturbed state. Therefore, VPIR can be given by $z^*=f_{\theta}(s^{\text{masked}}_t) = f_{\theta}(s^{\text{masked}+\text{aug}}_t)$. We use the auxiliary InfoNCE loss defined in Eq. \ref{equ:infonce} with the query key $\mathbf{q}= f_{\theta}(s^{\text{masked}}_t)$. The positive key is the representation extracted from the augmented-masked state $\mathbf{k}^+= f_{\theta}(s^{\text{masked}+\text{aug}}_t)$, and the negative key is extracted from other states at $t+1$ in the batch of samples. During the auxiliary update, the masker remains constant, and only the shared encoder $f_{\theta}$ is updated. This strategy mitigates the occurrence of non-stationary gradients originating from varied augmented RGB and stable depth inputs, thereby stabilizing the optimization procedure. Specifically, the masker receives updates exclusively from the environment's reward signal. VPIR is derived from a relatively stable masked visual distribution, resulting in improved sample efficiency and enhanced agent generalization capabilities.

\subsection{Learning Objectives}
\label{subsec:learning_objective}
The critic loss, $\mathcal{L}_Q$, is defined as a combination of Bellman residuals, as indicated in Eq. \ref{equ:bellman}, from $s^{\text{masked}}_t$ and its augmented counterpart $s^{\text{masked} + \text{aug}}_t$, as follows:
\begin{equation}
    \mathcal{L}_Q \left( s_t, a_t,r_t,s_{t+1} \right) = \alpha R^{2} +\beta  R^{2}_{\text{aug}},
    \label{eq:critic_loss}
\end{equation}
where $\alpha$ and $\beta$ are the coefficients to balance the ratio of the two data streams. 

The masker $M_{\theta}$ is updated along with the policy network by optimizing $\mathcal{L}_Q$ in Eq. \ref{eq:critic_loss}. In addition, a $\texttt{stop\_grad}$ operation is applied after $Q^{\text{aug}}$ estimation to avoid the impact of non-stationary gradients on Q-value estimation resulting from augmented observations. The target network $Q^{\text{tgt}}_\theta$ is updated using the exponential moving average, which is also isolated from $\mathcal{L}_{Q}$ by a $\texttt{stop\_grad}$ operation. It is noted that \textit{DeGuV} not only minimizes the distributional variance induced by data augmentation but also enhances VPIR learning through contrastive learning and stabilizes the optimization of $\mathcal{L}_Q$, thereby augmenting the agent's generalization capabilities and improving sampling efficiency. 

\section{Experiments \& Evaluations}
We evaluate our proposed approach in terms of generalization and data efficiency against three state-of-the-art RL algorithms focused on generalization: SGQN \cite{bertoin2022look}, SVEA \cite{hansen2021stabilizing}, and MaDi \cite{grooten2024madi}. As \textit{DeGuV} is based on DrQv2 \cite{yarats2021drqv2} and uses contrastive learning similar to CURL \cite{agarwal2021contrastive}, we also include these two algorithms for comparisons to see their respective improvements.

\subsection{Experiment Setup}
We evaluate our proposed approach in the \texttt{robosuite} simulation \cite{robosuite2020} using a virtual Franka Emika robot within the RL-Vigen benchmark \cite{yuan2024rl}. The experiments are conducted on four distinct tasks: \texttt{Lift}, \texttt{Door}, \texttt{NutAssemblyRound}, and \texttt{TwoArmPegInHole}. We use $\texttt{random\_shift}$ \cite{kostrikov2020image}, $\texttt{random\_overlay}$ \cite{hansen2021generalization} and $\texttt{random\_color\_jitter}$ \cite{ghiasi2021simple} augmentations: $\texttt{random\_shift}$ regularizes the encoder to prioritize salient features and enhance the agent's data efficiency, $\texttt{random\_overlay}$ interpolates between the observed image and a randomly selected image for randomizing distributions containing distractions, and $\texttt{random\_color\_jitter}$ creates the shifts of the contrast and hue color of observations.

We categorize the training and evaluation environments with difficulty level increases (\texttt{easy}, \texttt{medium}, \texttt{hard}). We train our models in a standard environment (\texttt{train}) and evaluate them in evaluation environments with diverse visual perturbations, including texture, color, and brightness alterations. Note that each model is trained on $1,000,000$ frame steps. For every $10,000$ frame steps, we re-evaluate the training policy for $10$ episodes with three random seeds.

\begin{table}[h]
    \vspace{-4pt}
    \centering
    \caption{Quantitative results on performance retention of \textit{DeGuV} compared to other baselines over different evaluation modes.}
    \label{tab:retaining}
    \begin{tabular}{c | cccccc} 
        \hline
        Mode & DrQv2 & CURL & SGQN & SVEA & MaDi & DeGuV \\ \hline \hline
        \texttt{easy} & $0.670$ & $\textbf{0.892}$ & $0.755$ & $0.830$ & $0.797$ & $0.873$ \\ 
        \texttt{medium} & $0.290$ & $0.441$ & $0.424$ & $0.227$ & $0.321$ & $\textbf{0.801}$ \\ 
        \texttt{hard} & $0.273$ & $0.343$ & $0.308$ & $0.236$ & $0.252$ & $\textbf{0.801}$ \\
        \hline
        \textbf{Average} & $0.411$ & $0.559$ & $0.495$ & $0.431$ & $0.457$ & $\textbf{0.825}$ \\ \hline
    \end{tabular}
    \vspace{-10pt}
\end{table}

\subsection{Generalization}

\begin{figure*}[t]
    \centering
    \vspace{4pt}
    \includegraphics[width=1.00\linewidth]{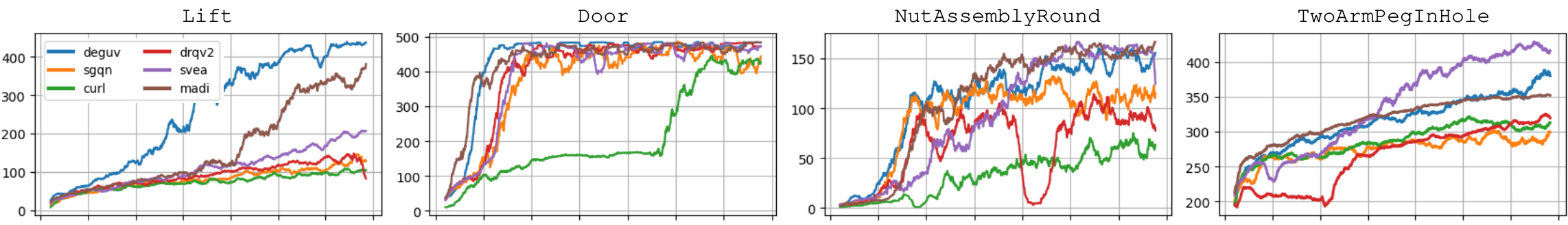}
    \caption{Comparisons of episode returns of \textit{DeGuV} compared to other baselines in four robotic tasks: \texttt{Lift}, \texttt{Door}, \texttt{NutAssemblyRound}, and \texttt{TwoArmPegInHole}, showing the our proposed method achieves better or similar data efficiency through the training process.}
    \label{fig:dataeff}
    \vspace{-15pt}
\end{figure*}

Table \ref{tab:eval_result} reports that \textit{DeGuV} showcases its generalization across all tasks and evaluation modes, achieving an average episode return of $311.44$, outperforming MaDi ($200.23$), SVEA ($196.24$), SGQN ($168.05$), CURL ($163.57$), and DrQv2 ($148.29$) in this scheme. Moreover, we find that \textit{DeGuV} retains $82.5\%$ of its training performance on average, far surpassing the baselines, as shown in Table \ref{tab:retaining}. From these two experiments, we conclude that the other algorithms are not able to adapt to the unseen visual distribution, while \textit{DeGuV} well manages this, thanks to depth-guided masking. Furthermore, our approach exhibits a significantly smaller performance drop in Table \ref{tab:retaining}, as the evaluation difficulty is more complex with visual perturbations, which highlights \textit{DeGuV}'s ability to address the generalization problem. Fig. \ref{fig:dataeff} additionally shows that \textit{DeGuV} achieves superior data efficiency on the \texttt{Lift} task and is competitive compared to the baselines on \texttt{Door}, \texttt{NutAssemblyRound}, and \texttt{TwoArmPegInHole} tasks.

\subsection{Interpretability}
\begin{figure}[t]
    \centering
    \includegraphics[width=0.93\linewidth]{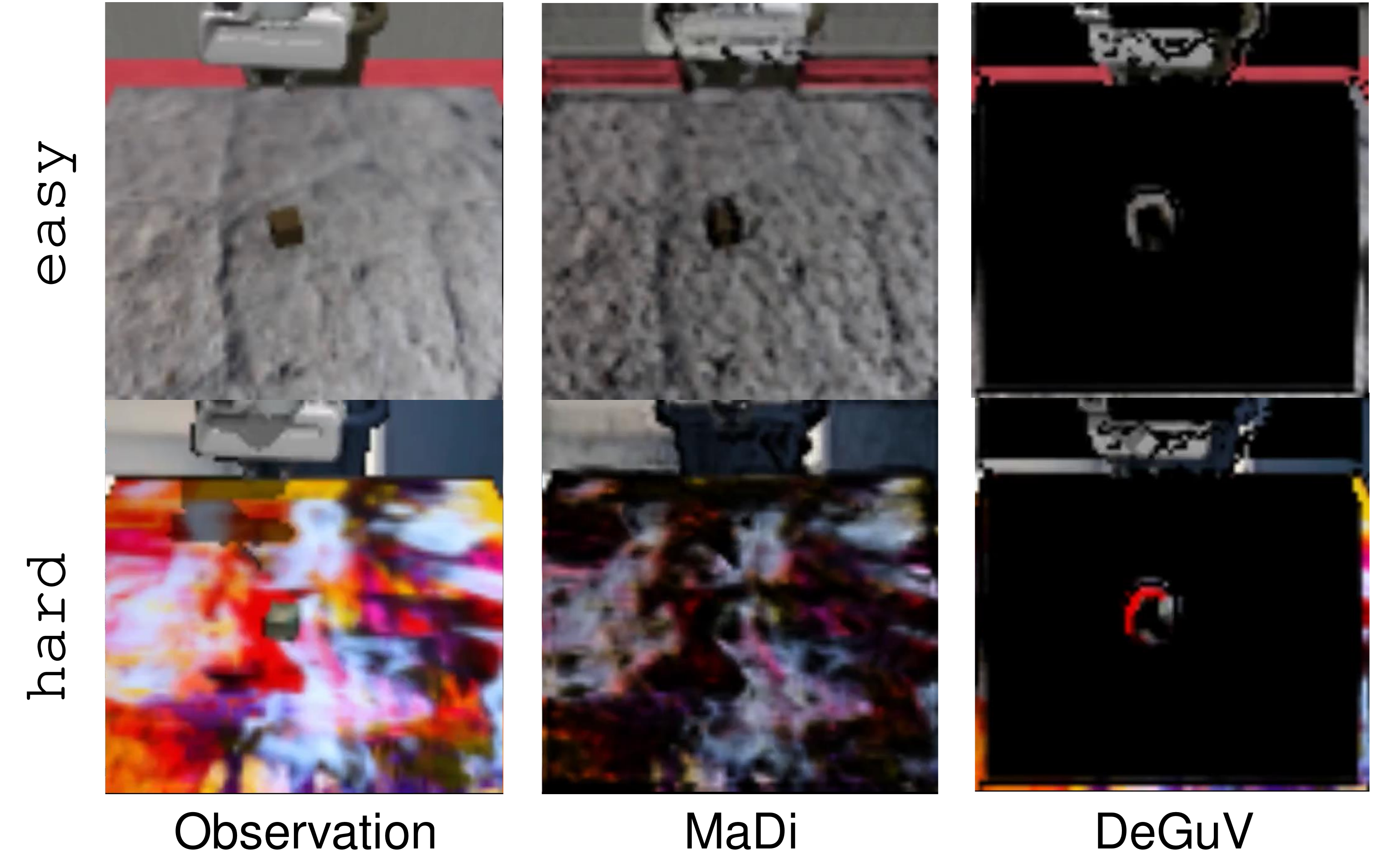}
    \caption{Qualitative results of masked observations between MaDi and \textit{DeGuV} from $\texttt{easy}$ and $\texttt{hard}$ of the \texttt{Lift} task.}
    \label{fig:mask_eval}
    \vspace{2pt}
\end{figure}

Given that the mask serves as a tool for interpreting the agent's decision-making rationale in response to an observation, we conduct analyses of the masks generated by \textit{DeGuV} and MaDi in evaluation modalities, as shown in Fig. \ref{fig:mask_eval}. Compared to MaDi, \textit{DeGuV} generates masks that remove most visual distractions, leaving approximately identical masked observations in the two evaluation modes. In this example, \textit{DeGuV} reveals merely $16.08\%$ of the visual observations in both modes, whereas MaDi reveals $76.24\%$ in $\texttt{easy}$ and $53.77\%$ in $\texttt{hard}$ mode. This enhances generalization capabilities in intricate visual environments and ensures performance retention.

\section{Sim-to-Real Transferability}
\label{sec:sim2real}

\begin{figure}[t]
    \centering
    \vspace{-4pt}
    \includegraphics[width=1.00\linewidth]{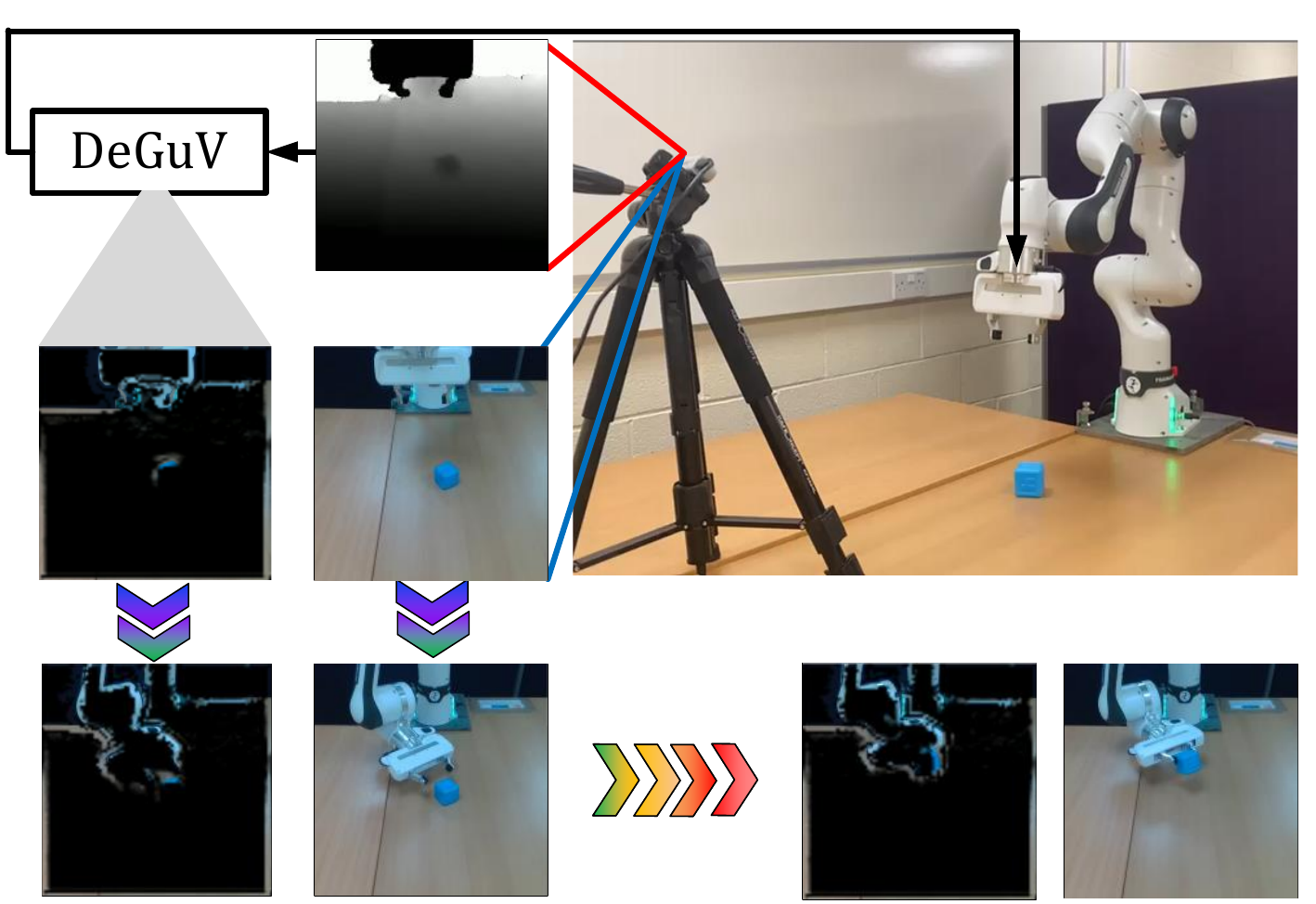}
    \caption{Experiment set-up and results of original and masked observations for the \texttt{Lift} task on a real-robot system.}
    \label{fig:real-world}
    \vspace{2pt}
\end{figure}

We deploy our trained model from \texttt{robosuite} to the real-world scenario to verify its sim-to-real transferability. The Franka Emika robot is set up with the \texttt{Lift} task. We wrap \textit{DeGuV} as a ROS2 package to interact with the robot via $\texttt{franka\_ros2}$. The ROS2 node processes images and controls the end-effector's delta positions and orientations, along with the gripper control. We then use $\texttt{panda}\text{-}\texttt{py}$ \cite{elsner2023taming} to perform desired robotic actions. An Intel D435i RealSense RGB-D camera is attached to acquire RGB-D observations. Inherent noises in depth frames are reduced by RealSense ROS spatial, temporal and hole-filling post-processing filters.

Fig. \ref{fig:real-world} shows that the masked observation produced by \textit{DeGuV} is similar to the masked observation in simulation, which only ``\textit{highlight}'' important objects and locations that are related to the task, such as the robot gripper, hand, and cube. This real-world experiment confirms the effectiveness of \textit{DeGuV} in generalization and interpretability. It also demonstrates the feasibility of sim-to-real transferability in a zero-shot manner. Our demonstration video is available at: \href{https://youtu.be/-Gt5i6Wi5Fs}{\small \texttt{https://youtu.be/-Gt5i6Wi5Fs}}.

\section{Conclusions and Future Works}

In this paper, we propose \textit{DeGuV}, targeting the generalization problem in visual reinforcement learning by leveraging the stable depth distribution to erase irrelevant pixels in the visual state. Our experiments show that the proposed method outperforms state-of-the-art algorithms on the RL-Vigen benchmark in generalizability. Our proposed method shows comparable sample efficiency during training and interpretability with masked observations compared to baselines. Finally, we demonstrate sim-to-real transferability, strengthening the generalization performance and potential for real robot deployment.

While depth input is invariant to visual perturbation, it usually contains noise, which is also worth investigating to improve \textit{DeGuV}'s performance in sim-to-real transfer. Furthermore, long-horizon complex tasks remain challenging for our model in visual RL models. In the future, we will conduct more real robot experiments to validate the performance of sim-to-real transferability and investigate its potential in more challenging and long-horizon tasks.

\bibliographystyle{IEEEtran}
\bibliography{IEEEabrv, 09_references}

\end{document}